\documentclass[letterpaper,journal]{IEEEtran}
\usepackage{amsmath,amsfonts}
\usepackage{algorithm} %
\usepackage{algpseudocode} %
\usepackage{array}
\usepackage[caption=false,font=normalsize,labelfont=sf,textfont=sf]{subfig}
\usepackage{textcomp}
\usepackage{stfloats}
\usepackage{url}
\usepackage{verbatim}
\usepackage{graphicx}
\usepackage{cite}

\usepackage[figuresright]{rotating}
\usepackage{multirow}
\usepackage{float}
\usepackage{booktabs}
\usepackage{threeparttable}
\usepackage{diagbox}
\usepackage[table]{xcolor}

\begin{document}

\title{Hierarchical Semantic-Constrained Heterogeneous Graph for
Audio-Visual Event Localization}

\author{Zhe Yang,
        Ruyi Zhang, 
        Hongtao Chen,
        Wenrui Li,~\IEEEmembership{~Member,~IEEE}
        Hengyu Man\\
        Wangmeng Zuo,~\IEEEmembership{~Senior Member,~IEEE}
        Xiaopeng Fan,~\IEEEmembership{~Senior Member,~IEEE}      
\thanks{This work was supported in part by the National Key R\&D Program of China (2023YFA1008501) and the National Natural Science Foundation of China (NSFC) under grant 624B2049 and U22B2035. (Corresponding author: Wenrui Li)}

\thanks{Zhe Yang, Ruyi Zhang, Hongtao Chen, Wenrui Li, Hengyu Man and Wangmeng Zuo are with the Faculty of Computing, Harbin Institute of Technology, Harbin 150001, China. (e-mail: yzhe610@stu.hit.edu.cn; 25B303031@stu.hit.edu.cn; ht166chen@163.com;  liwr@hit.edu.cn; manhengyu@hotmail.com; wmzuo@hit.edu.cn;).}

\thanks{Xiaopeng Fan is with the Faculty of Computing, Harbin Institute of Technology, Harbin 150001, China. He is also with the Peng Cheng Laboratory, Shenzhen 518000, China, and the Harbin Institute of Technology Suzhou Research Institute, Suzhou 215000, China. (e-mail:
fxp@hit.edu.cn).}}

\markboth{Journal of \LaTeX\ Class Files,~Vol.~14, No.~10, July~2024}%
{Shell \MakeLowercase{\textit{et al.}}: A Sample Article Using IEEEtran.cls for IEEE Journals}

\maketitle
\begin{abstract}
Open-vocabulary audio-visual event localization (OV-AVEL) jointly models audio-visual cues to recognize and temporally localize events, including categories unseen during training. Existing methods primarily learn joint audio-visual representations in Euclidean space, but still face two significant challenges. First, the lack of supervision signals for unseen categories makes it difficult to maintain audio-visual consistency across multiple temporal scales. Second, the lack of hierarchical constraints between segment- and video-level semantics prevents the model from establishing semantic consistency across different levels.
To address these challenges, we propose a hierarchical semantic constrained heterogeneous graph (HSCHG) for audio-visual event localization framework. We first construct a heterogeneous hierarchical graph in Euclidean space, which includes audio and visual segment nodes and their corresponding video-level nodes. We use multi-directional temporal edges to capture complete temporal information within each modality. Simultaneously, we employ a dual-threshold filtering gated fusion strategy, introducing cross-modal information only when the alignment confidence is high. Furthermore, we introduce bidirectional semantic constraints between segment- and video-level representations to achieve semantic consistency across different levels. Based on this, we map the multi-level audio-visual representations and text prototypes uniformly into hyperbolic space. We use a hierarchical entailment regularization loss to characterize the hierarchical relationships between videos and segments. Extensive experimental results show that our method outperforms existing methods on the OV-AVEL benchmark. Ablation studies further validate the effectiveness of our method.
\end{abstract}

\begin{IEEEkeywords}
Multi-modal learning,  event localization
\end{IEEEkeywords}

\section{Introduction}
\IEEEPARstart{W}{ith} the rapid growth of multimedia content, models need to understand multi-source information such as images, speech, and text simultaneously. Multi-modal learning has therefore become an important direction in video understanding, and has continuously made progress in tasks such as  image-text understanding \cite{MPARN,yang2,NSTRN,RCTRN}, video-text understanding \cite{song1,song2,yang1,yang3,song3}, and audio-visual learning \cite{STFT,MDST++,MDFT,MSTR,SpiVG}. The common goal of these tasks is to align the semantic and temporal cues of different modalities, utilizing complementary information to improve robustness and discriminative power. As a key task in complex video understanding, audio-visual event localization (AVEL) aims to jointly model audio and visual cues to identify event categories and locate their occurrence intervals. This capability is of great value in various applications, such as understanding and retrieving video content, short video recommendation, subtitle generation, and intelligent surveillance. AVEL can help systems automatically identify key event segments from noisy and complex videos, supporting more efficient content management.
\begin{figure}
    \centering    \includegraphics[width=0.9\linewidth]{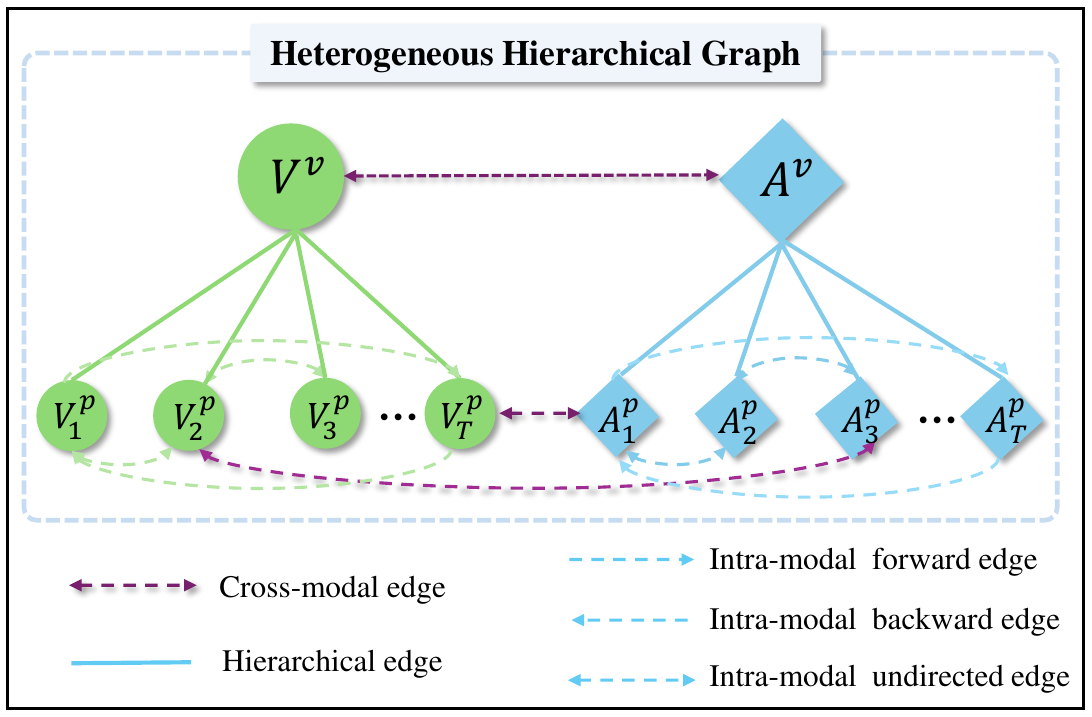}
    \caption{An illustration of the heterogeneous hierarchical graph in HSCHG. It contains segment-level visual nodes $\left \{V_t^p\right \} _{t=1}^T$ and audio nodes $\left \{A_t^p\right \} _{t=1}^T$, as well as video-level nodes $V^{v}$ and $A^{v}$. Intra-modal edges capture forward, backward, and undirected temporal dependencies. Hierarchical edges connect segments to video-level nodes for cross-granularity aggregation. Cross-modal edges link aligned audio and visual segments to learn audio-visual consistency under noise and misalignment.}
    \label{fig3}
\end{figure}
\begin{figure*}[t]
    \centering    \includegraphics[width=0.8\linewidth]{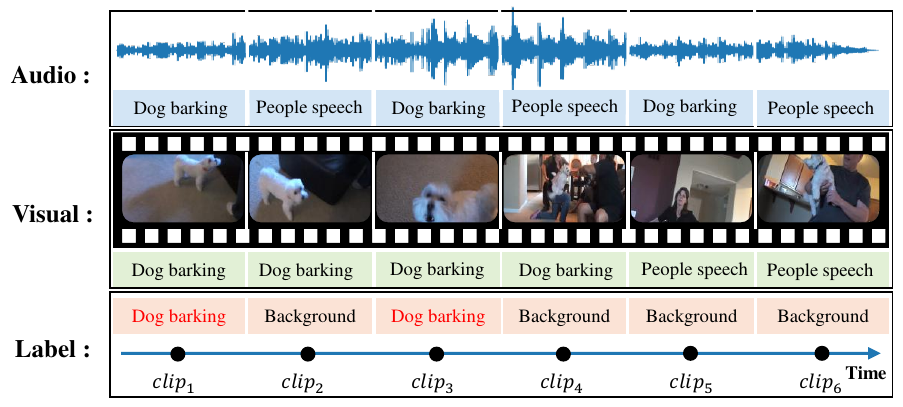}
    \caption{An illustration of OV-AVEL. The model jointly leverages segment-level audio and visual cues to determine whether a target event that is both audible and visible is present. This example highlights two key challenges in open-vocabulary settings: (1) asynchronous audio and visual signals can cause cross-modal semantic conflicts (e.g., in \textit{clip}$_2$, the audio indicates ``people speaking'' while the visual suggests ``dog barking''); (2) segments from previously seen categories may be mistakenly treated as evidence for the target event (e.g., in \textit{clip}$_6$, ``people speaking'' is incorrectly associated with the target event), resulting in false positives and inaccurate localization.}
    \label{fig1}
\end{figure*}

Most existing AVEL methods learn joint representations of audio and visual information in Euclidean space and perform inference at the video or segment level to accomplish event classification and temporal localization. Early work \cite{AVEL1,DCAN} defined AVEL as a temporal localization task that holds only when an event is simultaneously audible and visible, and established a basic framework using a sequence labeling paradigm. These methods typically assume that audio and visual cues are roughly synchronized in time. Therefore, they mainly rely on feature fusion or cross-modal attention to achieve better localization results. In practical scenarios, due to frequent camera changes and sudden noise in videos, the assumption of audio-visual synchronization often does not hold. Models easily mistake segments that are significant in only one modality but inconsistent across modalities as event evidence. CMBS \cite{AVEL2} re-models the task from the perspective of cross-modal background, defining the background as content with cross-modal semantic mismatch or appearing in only a single modality. To further improve the connection between different modes of data, the DMIN approach \cite{AVEL3} enhances cross-modal interaction by transitioning from sparse attention to a more concentrated form. It achieves this by incorporating a finer-grained audio-guided mechanism and a more comprehensive cross-channel attention model. Although the methods mentioned above improve robustness to noise and audio-visual inconsistencies, considering the background as a general category and learning a unified localization rule overlooks modality bias at the event level. Therefore, MAML \cite{AVEL4} proposes a localization paradigm that first predicts the event category and then performs segment localization based on the localization preference of that category. The above research mainly stays in a closed-set setting. If unseen events appear during the testing phase, the model can often only identify them as background or visible categories, making it difficult to provide clear semantic categories. Recently, works such as OV-AVE \cite{OVAVE,AVEL5} propose an open-vocabulary audio-visual event localization (OV-AVEL) task setting that is closer to real-world scenarios. This requires the model to simultaneously perform temporal localization of audible and visible events at the segment level and output specific category names for both seen and unseen categories.

Although the methods above have achieved promising progress, representation learning based on Euclidean space still faces critical bottlenecks in open-vocabulary scenarios. First, audio-visual events typically exhibit significant multi-scale variations along the time axis. The model needs to capture local segment information while achieving robust audio-visual synchronization across multi-scale temporal structures. Secondly, under open-vocabulary conditions, the lack of effective hierarchical constraints between segment semantics and video semantics prevents the model from establishing semantic consistency between videos and segments. As shown in Fig. \ref{fig1}, a trained model incorrectly identifies a segment containing "person speaking" (seen category) as the target event "dog barking." From a geometric representation perspective, the volume of Euclidean space grows only polynomially with the radius, making it difficult to effectively represent hierarchical relationships within modalities and tree-like structures between modalities. With its exponential volume growth property, hyperbolic space has a natural advantage in characterizing such hierarchical and tree-like structures \cite{PE}. Recent works \cite{hyperavca,shmamba} attempt to embed audio-visual features into a hyperbolic manifold and achieve modality alignment in the tangent space to improve zero-shot recognition performance. However, they only perform modality alignment on video-level representations, failing to effectively capture the hierarchical relationships between audio-visual temporal features and category semantics. Furthermore, the strong non-linear nature of hyperbolic space itself is not suitable for directly modeling complex intra-modal and inter-modal temporal dependencies. Therefore, the core of this research is to construct a robust multi-level temporal structure for audio-visual signals and model it jointly with category semantics. This joint modeling aims to improve generalization in open vocabulary audio-visual event localization by enforcing temporal consistency and semantic alignment across modalities and levels.

In this paper, we propose a hierarchical semantic-constrained heterogeneous graph (HSCHG) for audio-visual event localization. We first construct a heterogeneous hierarchical graph in Euclidean embedding space and jointly learn audio-visual consistency representations of segment-level and video-level semantics by aggregating features between cross-modal segment nodes and video-level nodes, as shown in Fig. \ref{fig3}. Subsequently, we map the multi-modal hierarchical features to hyperbolic space, using a hierarchical entailment loss to jointly constrain intra-modal and inter-modal relationships, ensuring consistency with the geometric structure of categorical semantic prototypes. Specifically, we construct a hierarchical heterogeneous graph using audio and visual segment-level nodes and their corresponding video-level nodes. For intra-modal segment nodes, we use undirected, forward, and backward edges to capture complete temporal information. During feature aggregation, we employ threshold-weighted aggregation to suppress the propagation of spurious neighbor information, thereby improving the robustness of temporal reasoning. For cross-modal nodes, we introduce a dual-threshold filtering and gated fusion mechanism, injecting cross-modal information only when the alignment confidence is high, thus reducing noise interference. Finally, we employ attention to establish bidirectional semantic constraints between segment-level and video-level representations. The segment sequence is calibrated for video-level semantic consistency, and the updated segment features refine the video-level semantics in reverse. Based on this multi-level representation, we uniformly map the video-level and segment-level audio-visual features and their corresponding hierarchical text embeddings into a hyperbolic manifold. We explicitly characterize the hierarchical relationships within and across modalities using an entailment cone loss based on hyperbolic geometry. Within modalities, we constrain the video-level embedding to contain its segment embeddings. Across modalities, we use text prototypes as parent nodes and impose entailment constraints on different levels of audio-visual embeddings, ensuring that multi-modal instances fall within the semantic cone of their corresponding categories and achieving hierarchical alignment.

Our main contributions are summarized as follows:
\begin{itemize}
\item We propose a hierarchical semantic-constrained heterogeneous graph framework for OV-AVEL. We jointly learn segment-level and video-level audio-visual consistency representations in Euclidean space, and further map multi-level instance and category semantics to hyperbolic space for geometric alignment, significantly improving generalization to unseen categories.

\item We design a heterogeneous hierarchical graph network. We also introduce multi-directional temporal edges and a gated fusion strategy with dual-threshold filtering to achieve robust cross-modal temporal reasoning.

\item We propose a hierarchical entailment regularization loss. Based on entailment cone constraints, we explicitly characterize the hierarchical relationships between videos and segments within modalities, and between text and audio-visual features across modalities.

\item Extensive experiments on the OV-AVEL benchmark dataset demonstrate that our method outperforms existing methods. Our ablation studies also validate the effectiveness of the proposed model and its components.
\end{itemize}

\section{Related Work}
\label{A}
\subsection{Audio-Visual Event Localization} 
audio-visual event localization requires jointly modeling audio and visual cues to locate audible and visible event segments on a timeline. Existing methods typically rely on feature fusion and temporal modeling, performing inference and scoring at the segment level. However, natural scenes often contain semantic and temporal inconsistencies caused by environmental noise and camera changes. Simple global fusion easily incorporates mismatched cross-modal information into event representations, reducing localization reliability.

To improve robustness, research emphasizes cross-modal consistency filtering and more refined interaction modeling. CSS-Net \cite{RD1} uses bidirectional guided collaborative attention to establish complementary relationships between audio and visual modalities. DMIN \cite{AVEL3} extends cross-modal interaction from sparse attention to dense cross-channel attention. It enhances correspondence modeling through finer-grained audio guidance and uses unimodal discriminative constraints to reduce the dilution of effective information during fusion. CCLN \cite{RD3} divides audio-visual integration into three stages: fusion, interaction, and integration. It introduces weakly supervised cross-modal contrastive constraints to make cross-modal semantics of the same category more concentrated, thus mitigating error amplification caused by misaligned samples. Besides consistency modeling, different events have varying dependencies on the two modalities, and a unified rule easily overlooks event-level modality bias. MAML \cite{AVEL4} adopts an event-aware localization paradigm that first predicts video event categories and then uses the category-specific modality preferences to guide segment localization. CCNet \cite{RD5} aggregates consistent event semantics using cross-modal attention. It introduces temporal consistency gating constraints to guide the other modality to focus on key segments in the temporal dimension and improves localization accuracy in overlapping scenes through multi-scale temporal collaboration. SVED \cite{RD6} uses multi-scale spatial attention to capture subtle motion differences. It enhances the completeness of fused representations by improving modality alignment and suppresses misleading irrelevant salient regions through spatial interference elimination. Recent works such as OV-AVE \cite{OVAVE} extend the task to an open-vocabulary setting, requiring the output of specific names and localization for unseen categories.

Although the above methods have made progress, there is still a lack of stable multi-level semantic constraints under open-vocabulary conditions to maintain consistency between video-level and segment-level semantics. From a multi-level modeling perspective, we construct a cross-modal hierarchical heterogeneous graph in Euclidean space to jointly learn segment-level and video-level audio-visual consistency representations, thereby improving audio-visual consistency under an open vocabulary setting.

\subsection{Hierarchical Representation Learning} 
Hierarchical representation learning focuses on organizing semantics at different granularities and establishing learnable cross-granularity constraints, enabling models to form stable global context while preserving local evidence. This problem is prevalent in data with compositional structures or conceptual hierarchies. For example, text consists of words and sentences, videos consist of frames and segments, and category semantics include superordinate and subordinate relationships. Effective hierarchical modeling typically requires addressing two points simultaneously: first, how to capture local dependencies and long-range relationships within a granularity, and second, how to establish consistency constraints between granularities to prevent local noise from disrupting global semantics.

In Euclidean space, mainstream methods often achieve cross-granularity aggregation through explicit hierarchical encoding structures, with different levels assuming different modeling responsibilities. HAN \cite{HAN} performs attention aggregation at the word and sentence levels, first forming sentence vectors and then aggregating them into document vectors, thus filtering local evidence at lower levels and obtaining more stable topic representations at higher levels. In video localization, HISAN \cite{HISAN} uses a two-level self-attention mechanism to model temporal dependencies and spatial context separately, allowing the segment level to focus on locally discriminative segments, and the global level to provide long-range context to support localization inference. In a weakly supervised setting, HiCo \cite{HiCo} decomposes consistency learning into local consistency over short time spans and thematic consistency over long time spans, mitigating the error accumulation caused by forcibly aligning all segments according to a uniform rule. From a geometric representation perspective, the volume of Euclidean space grows polynomially with the radius, making it difficult to accurately represent hierarchical relationships and tree structures with low distortion. Poincaré Embeddings \cite{PE} demonstrate that hyperbolic space can more effectively embed hierarchical data while preserving both similarity and hierarchical structure. HEC \cite{HEC} further explicitly parameterize partial order relationships using hyperbolic geodesic cones, giving entailment constraints a direct geometric interpretation. HyCoCLIP \cite{HyCoCLIP} organizes images, regions, and text descriptions into a cross-granularity hierarchical structure, jointly learning cross-granularity alignment with hyperbolic contrastive objectives and entailment cone constraints, thereby simultaneously improving hierarchical awareness and compositional generalization capabilities. For audio-visual Multi-modal data, Hyper-AVCA \cite{hyperavca} learns audio and visual embeddings in hyperbolic space and performs cross-modal alignment, demonstrating that hyperbolic geometry can serve as an effective representation carrier for Multi-modal hierarchical relationships.

Although the methods mentioned above have achieved promising results, OV-AVEL still faces higher demands for multi-level consistency. In this paper, we map the multi-level representations obtained through graph structures to hyperbolic space, explicitly modeling the hierarchical relationships between videos and segments, and between text and audio-visual modalities using implicit constraints, thereby improving the generalization ability of OV-AVEL.
\begin{figure*}
    \centering
    \includegraphics[width=1.0\linewidth]{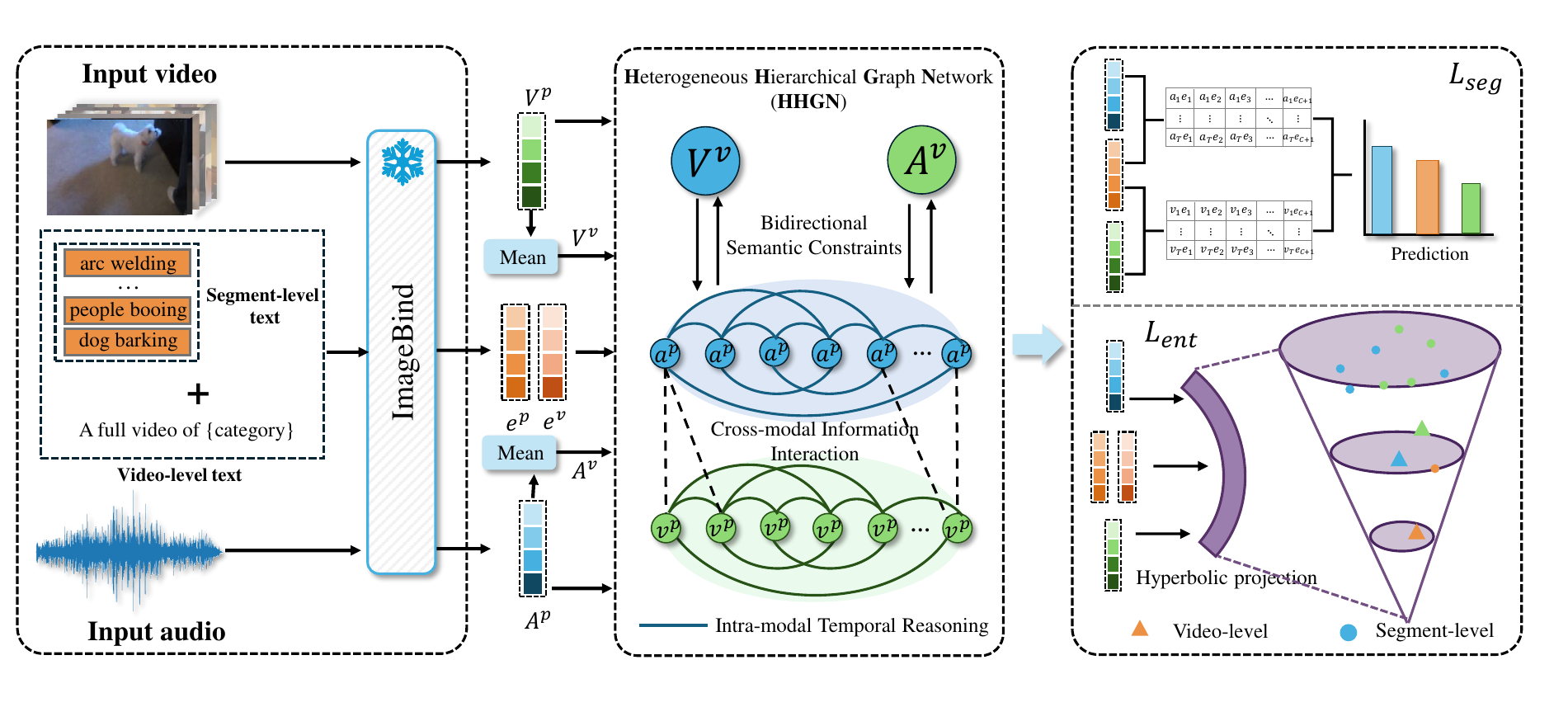}
    \caption{Architecture of HSCHG. We first extract features from audio, visual, and text inputs using a frozen pre-trained model. Subsequently, a heterogeneous hierarchical graph network performs temporal reasoning within each modality and facilitates information interaction between modalities, while simultaneously imposing bidirectional semantic constraints between segment-level and video-level representations. Finally, the model is optimized in hyperbolic space through hyperbolic projection, trained jointly using a segment localization loss $L_{seg}$ and a hyperbolic hierarchical entailment regularization term $L_{ent}$.}
    \label{fig2}
\end{figure*}

\section{Methodology}
\label{B}
\subsection{Task Definition} 
Given a video $X$, we split it into $T$ consecutive non-overlapping temporal segments and obtain segment-level audio-visual pairs $\{(a_t, v_t)\}_{t=1}^{T}$. In OV-AVEL, the test-time category space is open-vocabulary. The model is trained on seen classes but must generalize to both seen and unseen classes at test time. We denote the dataset as $D=\{(\mathbf{a}_i,\mathbf{v}_i,\mathbf{y}_i)\}$. For sample $i$, $\mathbf{a}_i=\{a_{i,t}\}_{t=1}^{T}$ and $\mathbf{v}_i=\{v_{i,t}\}_{t=1}^{T}$ are the audio and visual feature sequences, and $\mathbf{y}_i=\{y_{i,t}\}_{t=1}^{T}$ is the segment-level label sequence. The training split is $D_{\mathrm{train}}$ with label set $C_{\mathrm{seen}}$:
\begin{equation}
y_{i,t}\in\{0,1\}^{|C_{\mathrm{seen}}|+1}, \qquad 
\sum_{c\in C_{\mathrm{seen}}\cup\{0\}} y_{i,t,c}=1,
\end{equation}
where $c=0$ denotes the background class. The test split is $D_{\mathrm{test}}$ with an expanded label set:
\begin{equation}
C = C_{\mathrm{seen}}\cup C_{\mathrm{unseen}}, \qquad 
C_{\mathrm{seen}}\cap C_{\mathrm{unseen}}=\emptyset.
\end{equation}

The goal is to learn a classifier $f:(a_t,v_t)\mapsto p_t$, where $p_t\in\mathbb{R}^{|C|+1}$, that is trained with supervision from $C_{\mathrm{seen}}$ but can localize events over $C$ at test time.

The architecture of HSCHG is shown in Fig. \ref{fig2}. For fair comparison, we adopt the same pretrained feature extractor as OV-AVE \cite{OVAVE}. For each video, we obtain segment-level audio features $\mathbf{A}^p \in \mathbb{R}^{T\times D}$, segment-level visual features $\mathbf{V}^p \in \mathbb{R}^{T\times D}$, and category text features $\mathbf{E} \in \mathbb{R}^{(|C|+1)\times D}$. Temporal average pooling yields video-level features $\mathbf{A}^v,\mathbf{V}^v \in \mathbb{R}^{D}$. To build text prototypes aligned with video-level semantics for the entailment constraints, we also construct a video-level text feature for each sample. Given the event category, we use the prompt ``a full video of \{category\}'' and encode it with the same text encoder to obtain $e_v \in \mathbb{R}^{D}$. This $e_v$ is used to constrain consistency between video-level audio-visual representations and segment-level representations.

\subsection{Heterogeneous Graph Network}
To ensure semantic consistency between segments and the full video in event localization, we represent each video sample as a hierarchical heterogeneous graph. This model takes segment-level audio and visual sequence features as input and constructs a heterogeneous graph containing segment-level nodes and video-level nodes. For intra-modal segment nodes, we adopt threshold-weighted aggregation to suppress message propagation from background segments. For cross-modal nodes, we apply a dual-threshold filtering strategy combined with gated fusion to reduce noise diffusion caused by incorrect alignments. Finally, we impose bidirectional consistency constraints between the segment-level and video-level representations, thereby generating audio-visually consistent hierarchical representations. The node set of the graph consists of segment-level audio nodes $\{a_t\}_{t=1}^{T}$, segment-level visual nodes $\{v_t\}_{t=1}^{T}$, as well as a video-level audio node $a^v$ and a video-level visual node $v^v$. At layer $l$, the features of segment-level audio and visual nodes are denoted as $\{\mathbf{h}_t^{a,l}\}_{t=1}^{T}$ and $\{\mathbf{h}_t^{v,l}\}_{t=1}^{T}$, respectively. They are initialized as $\mathbf{h}_t^{a,(0)}=A_{:,t,:}^p$ and $\mathbf{h}_t^{v,(0)}=V_{:,t,:}^p$. The video-level representations at layer $l$ are denoted as $\mathbf{H}^{a,l}$ and $\mathbf{H}^{v,l}$.

\subsubsection{Intra-modal Temporal Reasoning}
To capture complete temporal dependencies, we construct multi-directional temporal edges (MDTE) within each modality. Following \cite{TPAMI}, for any segment indices $i,j \in \{1,\ldots,T\}$ within a temporal window $K_t$, we define three types of temporal edges as:
\begin{equation}
e_{ij}=\mathbb{I}\!\left(0<|i-j|\le K_t\right),
\end{equation}
\begin{equation}
e_{j\to i}=\mathbb{I}\!\left(0<i-j\le K_t\right),
\end{equation}
\begin{equation}
e_{i\to j}=\mathbb{I}\!\left(0<j-i\le K_t\right),
\end{equation}
where $e_{ij}=1$ indicates that an undirected edge exists between nodes $i$ and $j$. $e_{j\to i}=1$ indicates a forward directed edge from $j$ to $i$, and $e_{i\to j}=1$ indicates a backward directed edge from $i$ to $j$. Otherwise, the corresponding indicator equals $0$.

To suppress the propagation of background information, we adopt threshold-weighted neighbor aggregation. For an edge $i \rightarrow j$, we define node relevance using cosine similarity:
\begin{equation}
s_{ij}=\frac{\mathbf{h}_i^{\top}\mathbf{h}_j}{\|\mathbf{h}_i\|_2\|\mathbf{h}_j\|_2},
\label{eq9}
\end{equation}
where $\|\cdot\|_2$ denotes the $\ell_2$-norm. Given a intra-modal threshold $\tau$ and a weighting coefficient $\alpha_{ij}$, the intra-modal temporal feature aggregation is formulated as:
\begin{equation}
w_j(s_{ij})=\mathbb{I}(s_{ij}\ge\tau),
\end{equation}
\begin{equation}
\mathbf{m}_i=\sum_{j\in\mathcal{N}_{\mathrm{in}}(i)} \alpha_{ij}\,w_j(s_{ij})\,\mathbf{h}_j,
\end{equation}
where $\mathbb{I}(\cdot)$ is the indicator function and $\mathcal{N}_{\mathrm{in}}(i)$ denotes the neighbor set of node $i$.

We compute $\mathbf{m}^{a,l}_i$ and $\mathbf{m}^{v,l}_i$ over the three temporal relations in the audio and visual modalities, respectively. Then, we update segment-node features using a feed-forward network:
\begin{equation}
\widetilde{\mathbf{h}}^{x,l}_t=\mathrm{LN}\big(\mathbf{h}^{x,l}_t+\mathrm{FFN}(\mathbf{m}^{x,l}_t)\big),\quad x\in\{a,v\},
\end{equation}
where $\mathrm{FFN}(\cdot)$ is a feed-forward network composed of two linear layers and $\mathrm{GELU}(\cdot)$, and $\mathrm{LN}(\cdot)$ denotes layer normalization.

\subsubsection{Cross-modal Information Interaction}
Cross-modal information is heterogeneous, and directly fusing all cross-modal neighbors can easily introduce noise due to alignment errors. We introduce a dual-threshold gating mechanism (DTGM) to filter and reweight cross-modal messages. Taking an audio node $a_i$ as an example, given an alignment radius $P$, we establish a cross-modal edge between $a_i$ and $v_j$ when $|i-j|\le P$. Similarly, we compute the cross-modal node relevance $s_{ij}$ using the same cosine similarity as in Eq.(\ref{eq9}). Given two cross-modal thresholds $\tau_1$ and $\tau_2$, the cross-modal piecewise weight is defined as:
\begin{equation}
w_j(s_{ij})=
\begin{cases}
0, & s_{ij}<\tau_1,\\
w_1, & \tau_1\le s_{ij}<\tau_2,\\
w_2, & s_{ij}\ge\tau_2,
\end{cases}
\end{equation}
where $w_1$ and $w_2$ are predefined prior weights. This design makes cross-modal propagation more conservative in early training, thereby improving optimization stability. The cross-modal aggregation for audio node $a_i$ is:
\begin{equation}
\mathbf{m}^{a,l}_{i}=\sum_{j\in\mathcal{N}_{\mathrm{cr}}(a_i)} \alpha_{j}\, w_{j}\!\left(s_{ij}\right)\widetilde{\mathbf{h}}^{v,l}_{j},
\end{equation}
where $\mathcal{N}_{\mathrm{cr}}(a_i)$ is the set of visual neighbors connected to node $a_i$. To prevent cross-modal messages from directly overriding intra-modal semantics under low-confidence scenarios, we introduce a gating vector:
\begin{equation}
\mathbf{g}^{a,l}_{t}=\sigma\!\left(\mathbf{W}^{a}_{g}\big[\widetilde{\mathbf{h}}^{a,l}_{t}\ \big|\ \mathbf{m}^{a,l}_{t}\big]\right),
\end{equation}
where $[\cdot|\cdot]$ denotes vector concatenation, $\mathbf{W}^{a}_{g} \in \mathbb{R}^{C\times 2C}$ are learnable parameters, and $\sigma(\cdot)$ is the Sigmoid function.

The updated audio-node feature is then computed as:
\begin{equation}
\mathbf{h}^{a,l+1}_{t}=\mathrm{LN}\!\left(\widetilde{\mathbf{h}}^{a,l}_{t}+\lambda_l\left(\mathbf{g}^{a,l}_{t}\odot \mathbf{m}^{a,l}_{t}\right)\right),
\end{equation}
where $\odot$ denotes element-wise multiplication and $\lambda_l$ is a layer scaling coefficient. The update method for the visual node features $\mathbf{h}^{v,l+1}_{t}$ is similar.

\subsubsection{Bidirectional Semantic Constraints}
Segment-level representations are used for interval discrimination, emphasizing whether a segment matches the target event, but they are susceptible to shot changes and non-target sounds. Video-level representations provide global contextual constraints and are better suited to capturing overall event semantics, yet they cannot directly yield precise temporal boundaries. Motivated by their complementarity, we introduce bidirectional semantic constraints (BSC). Video-level semantics calibrate segment responses in a top-down manner, while segment evidence refines video-level representations in a bottom-up manner, forming a closed-loop consistency.

We first perform top-down semantic calibration. We map the video-level semantics of both modalities into context vectors:
\begin{equation}
\mathbf{c}^{a,l}=\mathbf{W}_c^a\mathbf{H}^{a,l}+\mathbf{W}_c^a\mathbf{H}^{v,l},
\end{equation}
\begin{equation}
\mathbf{c}^{v,l}=\mathbf{W}_c^v\mathbf{H}^{v,l}+\mathbf{W}_c^a\mathbf{H}^{a,l},
\end{equation}
and then calibrate segment representations as:
\begin{equation}
\overline{\mathbf{h}}_t^{a,l} =\mathrm{LN}(\mathbf{h}_t^{a,l}+\gamma\mathbf{c}^{a,l}),
\end{equation}
\begin{equation}
\overline{\mathbf{h}}_t^{v,l} =\mathrm{LN}(\mathbf{h}_t^{v,l}+\gamma\mathbf{c}^{v,l}),
\end{equation}
where $\gamma$ is a learnable intensity coefficient used to control the magnitude of the influence of video-level context on the response of local segments. $\overline{\mathbf{h}}_t^{a,l}$ and $\overline{\mathbf{h}}_t^{v,l}$ are the calibrated segment-level representations, and $\mathrm{LN}(\cdot)$ denotes layer normalization. This process reduces the interference of local noise on localization by imposing global semantic constraints. Next, we refine video-level semantics using segment-level features:
\begin{equation}
p^{a,l}_{t}=\frac{\exp\!\left(\mathbf{w}^{a\top}\overline{\mathbf{h}}^{a,l+1}_{t}\right)}{\sum_{k=1}^{T}\exp\!\left(\mathbf{w}^{a\top}\overline{\mathbf{h}}^{a,l+1}_{k}\right)},
\end{equation}
\begin{equation}
\mathbf{H}^{a,l+1}=\mathrm{LN}\!\left(\mathbf{H}^{a,l}+\sum_{t=1}^{T} p^{a,l}_{t}\,\mathbf{W}^{a}_{p}\overline{\mathbf{h}}^{a,l+1}_{t}\right),
\end{equation}
where $\mathbf{w}^{a} \in \mathbb{R}^C$ and $\mathbf{W}^{a}_{p} \in \mathbb{R}^{C \times C}$ are the attention scoring vector and projection matrix, respectively. The visual branch obtains $\mathbf{H}^{v,l+1}$ in the same way. This bottom-up aggregation makes the video-level representation focus on the accumulation of event-relevant segments, thereby providing a more reliable semantic prior and improving localization robustness.

\subsection{Loss Function}
We learn localizable audio-visual representations via segment-level classification supervision, and further introduce hierarchical entailment regularization in hyperbolic space to explicitly enforce semantic consistency between video-level and segment-level representations. The overall optimization objective is:
\begin{equation}
\mathcal{L} = \mathcal{L}_{\mathrm{seg}} + \lambda \mathcal{L}_{\mathrm{ent}},
\end{equation}
where $\mathcal{L}_{\mathrm{seg}}$ is supervised using segment annotations of seen classes in the training set. $\mathcal{L}_{\mathrm{ent}}$ does not require extra annotations. Instead, it stabilizes cross-modal alignment under open-vocabulary testing by imposing hierarchical partial-order constraints. $\lambda$ is a weighting coefficient.

\subsubsection{Cross-Entropy Classification Loss}
To ensure a fair comparison, we adopt the same segment-level open-vocabulary event classification supervision as OV-AVE \cite{OVAVE}, using $\mathcal{L}_{\mathrm{seg}}$. Specifically, we compute cosine similarity matrices between the audio and visual segment features produced by the heterogeneous hierarchical graph network and the class text feature matrix $\mathbf{E}$. Let the segment-level audio and visual node outputs from the last graph layer be $\{\mathbf{g}_{p,t}^{a}\}_{t=1}^{T}$ and $\{\mathbf{g}_{p,t}^{v}\}_{t=1}^{T}$. The audio--text and visual--text similarity matrices are defined as:
\begin{equation}
\mathbf{S}_{at}= \big\|\mathbf{G}^a\big\|_2 \big\|\mathbf{E}\big\|_2^{\top} \in \mathbb{R}^{T\times (|C|+1)},
\end{equation}
\begin{equation}
\mathbf{S}_{vt}= \big\|\mathbf{G}^v\big\|_2 \big\|\mathbf{E}\big\|_2^{\top} \in \mathbb{R}^{T\times (|C|+1)},
\end{equation}
where $\mathbf{G}^a=[\mathbf{g}_1^a;\ldots;\mathbf{g}_T^a] \in \mathbb{R}^{T \times D}$, $\mathbf{G}^v=[\mathbf{g}_1^v;\ldots;\mathbf{g}_T^v] \in \mathbb{R}^{T \times D}$, and $\|\cdot\|_2$ denotes $L_2$ normalization along the last dimension. Since an event occurrence should be supported by both audio and visual cues, we emphasize consistent responses across modalities via element-wise multiplication and obtain the segment-level class distribution:
\begin{equation}
\mathbf{P}= \mathrm{Softmax}\!\left( \sqrt{\mathrm{ReLU}\!\left(\mathbf{S}_{at} \odot \mathbf{S}_{vt}\right)} \right),
\end{equation}
where $\odot$ is the Hadamard product and $\mathbf{P} \in \mathbb{R}^{T\times (|C|+1)}$. Given segment labels $y^{s}_{i,t}\in\{0,1\}^{|C_{\mathrm{seen}}|+1}$, we construct supervision using only seen classes during training:
\begin{equation}
L_{\mathrm{seg}}=\frac{1}{NT} \sum_{i=1}^N \sum_{t=1}^T \mathrm{CE}\!\left(\mathbf{P}_{i,t,:},\,y^{s}_{i,t}\right).
\end{equation}

\subsubsection{Hierarchical Entailment Regularization Loss}
Under the open-vocabulary setting, training supervision covers only $C_{\mathrm{seen}}$, while test-time inference requires aligning segment representations with a larger text category space. To introduce stable semantic constraints without additional annotations, we construct entailment-cone constraints in the Lorentz-model hyperbolic space $\mathbb{H}^{D}_{c}$ and regularize event segments. Let $\mathbf{g}^{a}_{\mathrm{v}}$ and $\mathbf{g}^{v}_{\mathrm{v}}$ denote the video-level audio and visual representations, and let $\mathbf{e}_t$ and $\mathbf{e}_{\mathrm{v}}$ denote the corresponding segment-level class text feature and video-level class text feature, respectively. We map them into $\mathbb{H}^{D}_{c}$ via a linear layer $\mathbf{W} \in \mathbb{R}^{D \times D}$ and a hyperbolic projection $\Phi_{c}(\cdot)$\cite{HyCoCLIP}:
\begin{equation}
\begin{aligned}
\mathbf{z}^{a}_{p,t} \; &=\; \Phi_{c}\!\left(\mathbf{W}\mathbf{g}^{a}_{p,t}\right), 
&\qquad
\mathbf{z}^{v}_{p,t} \; &=\; \Phi_{c}\!\left(\mathbf{W}\mathbf{g}^{v}_{p,t}\right), \\[2pt]
\mathbf{z}^{a}_{\mathrm{v}} \; &=\; \Phi_{c}\!\left(\mathbf{W}\mathbf{g}^{a}_{\mathrm{v}}\right), 
&\qquad
\mathbf{z}^{v}_{\mathrm{v}} \; &=\; \Phi_{c}\!\left(\mathbf{W}\mathbf{g}^{v}_{\mathrm{v}}\right), \\[2pt]
\mathbf{e}^{\mathbb{H}}_{t} \; &=\; \Phi_{c}\!\left(\mathbf{W}\mathbf{e}_{t}\right), 
&\qquad
\mathbf{e}^{\mathbb{H}}_{\mathrm{v}} \; &=\; \Phi_{c}\!\left(\mathbf{W}\mathbf{e}_{\mathrm{v}}\right).
\end{aligned}
\end{equation}

Given $\mathbf{x},\mathbf{y}\in\mathbb{H}^{D}_{c}$, where $\mathbf{x}$ represents a more general parent concept, we use a hinge-style entailment cost to measure whether $\mathbf{y}$ lies inside the semantic cone oriented by $\mathbf{x}$:
\begin{equation}
h(\mathbf{x},\mathbf{y})=\max\big(0,\ \angle(\mathbf{x},\mathbf{y})-\mathrm{ap}(\mathbf{x})\big),
\end{equation}
where $\angle(\mathbf{x},\mathbf{y})$ is the oxy-angle in the Lorentz model and $\mathrm{ap}(\mathbf{x})$ is the half-aperture \cite{Lorentz,HALF}. We construct regularization terms from three types of relationships. First, the video-level representation characterizes the event context of the entire video, while the segment-level representation describes its instantiation on the timeline. Therefore, we constrain the segment embedding to be contained within the corresponding video embedding:
\begin{equation}
\mathcal{L}_{\mathrm{in}}
=
\frac{1}{|\mathcal{T}^{+}|}\sum_{t\in\mathcal{T}^{+}}
\Big(
h(\mathbf{z}^{a}_{\mathrm{v}},\mathbf{z}^{a}_{p,t})
+
h(\mathbf{z}^{v}_{\mathrm{v}},\mathbf{z}^{v}_{p,t})
\Big),
\end{equation}
where $\mathcal{T}^{+}$ denotes the set of segments in the video that belong to the event:
\begin{equation}
\mathcal{T}^{+}=\left\{\, t \in \{1,\ldots,T\}\ \big|\ y^{s}_{i,t}=1 \,\right\},
\end{equation}


Visual and audio observations do not exactly align with textual descriptions. Textual semantics are typically more abstract and exhibit superordinate conceptual properties, making them more general than audio and visual representations \cite{HyCoCLIP}. We treat text as higher-level parent concepts and enforce video-level and segment-level embeddings to be semantically entailed by their corresponding textual embeddings:
\begin{equation}
\begin{aligned}
\mathcal{L}_{\mathrm{cr}}
&=
h\!\left(\mathbf{e}^{\mathbb{H}}_{\mathrm{v}},\mathbf{z}^{a}_{\mathrm{v}}\right)
+
h\!\left(\mathbf{e}^{\mathbb{H}}_{\mathrm{v}},\mathbf{z}^{v}_{\mathrm{v}}\right)
\\
&\quad
+
\frac{1}{|\mathcal{T}^{+}|}\sum_{t\in\mathcal{T}^{+}}
\Big(
h\!\left(\mathbf{e}^{\mathbb{H}}_{t},\mathbf{z}^{a}_{p,t}\right)
+
h\!\left(\mathbf{e}^{\mathbb{H}}_{t},\mathbf{z}^{v}_{p,t}\right)
\Big).
\end{aligned}
\end{equation}

The hierarchical entailment regularization loss is defined as:
\begin{equation}
\mathcal{L}_{\mathrm{ent}}=\mathcal{L}_{\mathrm{in}} + \mathcal{L}_{\mathrm{cr}}.
\end{equation}

\begin{table*}
    \centering
    \begin{threeparttable}
        \caption{The performance comparison on OV-AVEBench datasets.}
        \label{MAIN}
        \fontsize{10}{14}\selectfont
        \setlength{\tabcolsep}{3.5pt}
        \begin{tabular}{c|cccc|cccc|cccc}
            \toprule[1.5pt]
            \multirow{2}{*}{Model} &
            \multicolumn{4}{c|}{Seen} &
            \multicolumn{4}{c|}{Unseen}&
            \multicolumn{4}{c}{Total} \\
             & $Acc.$ &$ Seg.$ & $Eve.$ & \textbf{$Avg.$} & $Acc.$ & $Seg.$ & $Eve.$ & \textbf{$Avg.$} & $Acc.$ & $Seg.$ & $Eve.$ & \textbf{$Avg.$}  \\
            \midrule
            CMRA \cite{CMRA}
            &65.2 &58.8 &54.3 &59.4 &36.0 &31.0 &26.3 &31.3 &44.3 &38.9 &34.3 &39.2\\
            AVE\cite{AVE} 
            &76.6 &63.6 &56.0 &65.4 &44.6 &33.2 &24.0 &34.0 &53.8 &41.9 &33.2 &42.9\\
            PSP\cite{PSP} 
            &75.4 &66.8 &61.0 &67.7 &33.7 &28.2 &24.2 &28.7 &45.6 &39.3 &34.7 &39.9\\
            MM-Pyramid \cite{MM}
            &\textbf{76.5} &\textbf{66.9} &\underline{62.3} &\underline{68.6} &36.8 &29.0 &23.8 &29.9 &48.4 &40.2 &35.2 &41.2\\
            OV-AVE\cite{OVAVE}
            &72.5 &61.8 &54.5 &62.9 &\underline{64.9} &\underline{55.0} &\underline{47.5} &\underline{55.8} &\underline{67.1} &\underline{56.9} &\underline{49.5} &\underline{57.8}\\
            \cline{1-13}
            \textbf{HSCHG} & \underline{76.4} & \underline{66.8} & \textbf{63.1} & \textbf{68.8} & \textbf{65.6} & \textbf{55.7} &\textbf{48.2} &\textbf{56.5} &\textbf{68.5} &\textbf{58.8} &\textbf{51.6} &\textbf{59.7}\\
            \bottomrule[1.5pt]
        \end{tabular}
    \end{threeparttable}
\end{table*}

\section{Experiments}
\label{C}

\subsection{Evaluation Metrics and Dataset}
We primarily conduct OV-AVEL experiments on the OV-AVEBench dataset\cite{OVAVE}, and evaluate fine-tuning baseline schemes on this benchmark. To ensure the reproducibility of the experiments and a fair comparison, we strictly follow the baseline protocol. Specifically, we use the same seen and unseen category definitions, as well as the same training, validation, and test splits. We also adopt the same evaluation metrics, including $Acc.$, $Seg.$, $Eve.$, and their mean $Avg.$.

OV-AVEBench is built based on YouTube video resources from VGGSound. From VGGSound's 309 categories, highly similar categories and those rarely seen in real-world scenarios were filtered out, ultimately retaining 67 common event categories more suitable for judging audio-visual semantic consistency. These cover human activities, animal activities, musical instruments, transportation, and other scenarios. Each video is evenly divided into 10 one-second segments, with the middle frame of each segment representing the visual content and the corresponding one-second audio representing the auditory content. When the audio and visual semantics are consistent, the segment is labeled as a positive sample. Otherwise, it is labeled as the background category. Under open-vocabulary segmentation, 46 of the 67 categories are visible training categories, while the remaining 21 categories only appear during the evaluation phase. The ratio of videos with visible training categories to those without in the validation and test sets is approximately 3:7 to systematically evaluate the model's generalization ability to unseen categories.

\subsection{Experimental Settings}
To ensure consistency with prior work, we follow the same feature processing pipeline and training strategy as the baseline. We adopt ImageBind \cite{image} as the feature extractor to obtain 1024-dimensional audio, visual, and text representations, respectively. The number of Temporal Layers after feature extraction is kept unchanged at one. The model is trained on a single 40 GB NVIDIA A100 GPU. For optimization, we use the Adam optimizer with an initial learning rate of \(5 \times 10^{-5}\). The batch size is set to 32, and training is conducted for 10 epochs. In the quantitative analysis, we investigate the effects of different hyperparameters in the proposed method. Specifically, the final hyperparameter settings are as follows. The intra-modal threshold is set to \(\tau = 0.5\). The cross-modal thresholds are set to \(\tau_1 = 0.2\) and \(\tau_2 = 0.5\). The intensity coefficient is set to \(\gamma = 0.1\). The entailment-cone loss weight is set to \(\lambda = 0.1\).

\begin{table*}
    \centering
    \begin{threeparttable}
        \caption{Ablation for HSCHG on OV-AVEBench datasets}
        \label{ABLA}
        \fontsize{10}{14}\selectfont
        \setlength{\tabcolsep}{3.5pt}
        \begin{tabular}{c|cccc|cccc|cccc}
            \toprule[1.5pt]
            \multirow{2}{*}{\textbf{Method}} &
            \multicolumn{4}{c|}{\textbf{Seen}} &
            \multicolumn{4}{c|}{\textbf{Unseen}}&
            \multicolumn{4}{c}{\textbf{Total}} \\
             & $Acc.$ & $Seg.$ & $Eve.$ & \textbf{$Avg.$} & $Acc.$ & $Seg.$ & $Eve.$ & \textbf{$Avg.$} & $Acc.$ & $Seg.$ & $Eve.$ & \textbf{$Avg.$}  \\
            \midrule
            baseline
           &72.5 &61.8 &54.5 &62.9 &64.9 &55.0 &47.5 &55.8 &67.1 &56.9 &49.5 &57.8\\
            W/o HHGN 
            &73.2 &64.7 &56.9 &64.9 &65.1 &56.0 &47.6 &56.2 &67.6 &57.9 &50.2 &58.6\\
            W/o $\mathcal{L}_{ent}$
            &74.6 &62.9 &59.7 &65.7 &65.3 &55.4 &48.5 &56.4 &68.0 &57.2 &51.0 &58.7\\
            \cline{1-13}
            \textbf{HSCHG} & 76.4  & 66.8  & 63.1  & \textbf{68.8}  & 65.6  & 55.7  &48.2 &\textbf{56.5} &68.5 &58.8 &51.6 &\textbf{59.7}\\
            \bottomrule[1.5pt]
        \end{tabular}
    \end{threeparttable}
\end{table*}

\begin{table}
	\centering
	\begin{threeparttable}
		\fontsize{9}{13}\selectfont 
		\setlength{\tabcolsep}{11pt} 
		\begin{tabular}{c|ccc|c}  
			\toprule[1.5pt]  
			\textbf{Model}  & $Acc.$ &$ Seg.$ & $Eve.$ & \textbf{$Avg.$} \\ 
			\hline
 			baseline   & 67.1 & 56.9 & 49.5 & 57.8 \\
 			W/o MDTE   & 67.7 & 57.5 & 50.8 & 58.5 \\
 			W/o DTGM   & 67.4 & 57.8 & 50.6 & 58.8 \\
 			W/o BSC  & 68.2 & 58.3 & 50.1  & 59.3 \\
            \hline
  			HHGN  & \textbf{68.5} & \textbf{58.8} & \textbf{51.6} & \textbf{59.7} \\
			\bottomrule[1.5pt]
		\end{tabular}
        		\caption{Ablation study of the HHGN.}
                		\label{HHGN}
	\end{threeparttable}
\end{table}

\subsection{Main Results and Comparison}
We compare HSCHG with CMRA \cite{CMRA}, AVE \cite{AVE}, PSP \cite{PSP}, MM-Pyramid \cite{MM}, and the OV-AVE \cite{OVAVE}. Table \ref{MAIN} reports $Acc.$, $Seg.$, $Eve.$, and their mean $Avg.$ for HSCHG and representative method under three evaluation settings: seen, unseen, and total categories. Performance differences are modest on seen categories but widen substantially on unseen categories, indicating that methods that jointly maintain localization quality and open-vocabulary generalization are most competitive. On seen categories, HSCHG outperforms OV-AVE, raising the $Avg.$ from 62.9 to 68.8. This demonstrates that the proposed method performs excellently in open-vocabulary scenarios while remaining effective in closed-set scenarios. Compared with MM-Pyramid, HSCHG is slightly lower on $Acc.$ and $Seg.$ but higher on $Eve.$, which leads to a better $Avg.$. This phenomenon indicates that HSCHG does not over-optimize metrics at the segment level. Instead, it enhances the aggregation and differentiation capabilities of event-level evidence. On unseen categories, several methods with a bias towards closed-set settings showed significant performance degradation. Thanks to the structured prior provided by the hierarchical entailment loss introduced in HSCHG, the model maintains stronger generalization ability and achieves superior performance on unseen categories. In the all-category setting, HSCHG achieved an $Avg.$ of 59.7, significantly higher than OV-AVE's 57.8, further demonstrating the method's stable gains and stronger robustness under different category divisions. These improvements arise from hierarchical segment- and video-level semantic constraints that suppress the influence of noisy segments on global semantics, together with non-Euclidean modeling of hierarchical entailment that provides stable structured priors for unseen categories. Consequently, HSCHG improves robustness and cross-category generalization for open-vocabulary localization without additional supervision.

\begin{table}
	\centering
	\begin{threeparttable}
		\fontsize{9}{13}\selectfont 
		\setlength{\tabcolsep}{11pt} 
		\begin{tabular}{c|ccc|c}  
			\toprule[1.5pt]  
			\textbf{Loss}  & $Acc.$ & $Seg.$ & $Eve.$ & \textbf{$Avg.$} \\ 
			\hline
 			$\mathcal{L}_{seg}$   & 68.0 & 57.2 & 51.0 & 58.7 \\
 			$\mathcal{L}_{seg}$ + $\mathcal{L}_{in}$   & 68.1 & 57.8 & 51.1 & 59.0 \\
 			$\mathcal{L}_{seg}$ + $\mathcal{L}_{cr}$   & 68.2 & 58.2 & 50.8  & 59.1 \\
            \hline
  			$\mathcal{L}_{seg}$ + $\mathcal{L}_{ent}$  & \textbf{68.5} & \textbf{58.8} & \textbf{51.6} & \textbf{59.7} \\
			\bottomrule[1.5pt]
		\end{tabular}
        		\caption{Ablation study of the loss function.}
                		\label{LOSS}
	\end{threeparttable}
\end{table}

\subsection{Ablation Study}
This section investigates the contribution of each key component within the proposed HSCHG framework.

\subsubsection{Effectiveness of HSCHG Components}
Table \ref{ABLA} evaluates the impact of the Heterogeneous Hierarchical Graph Network (HHGN) and the Hierarchical Entailment Regularization Loss ($\mathcal{L}_{ent}$). On seen categories, removing HHGN causes a significant performance drop (Avg. declines from 68.8 to 64.9), confirming the necessity of the graph structure for learning robust semantic consistency between segments and video-level context. Furthermore, the exclusion of $\mathcal{L}_{ent}$ degrades performance across all settings. Specifically, the Avg. on unseen categories drops, indicating that $\mathcal{L}_{ent}$ provides essential structured priors for open-vocabulary generalization. By enforcing entailment constraints where video-level representations encompass segment-level evidence and align with text prototypes, $\mathcal{L}_{ent}$ ensures geometric consistency in the hyperbolic space, which is critical for handling unseen categories.

\begin{figure}[t]
  \centering
  \captionsetup[subfigure]{labelformat=empty}
  \subfloat[(a). $\gamma$]{\includegraphics[width=0.48\linewidth]{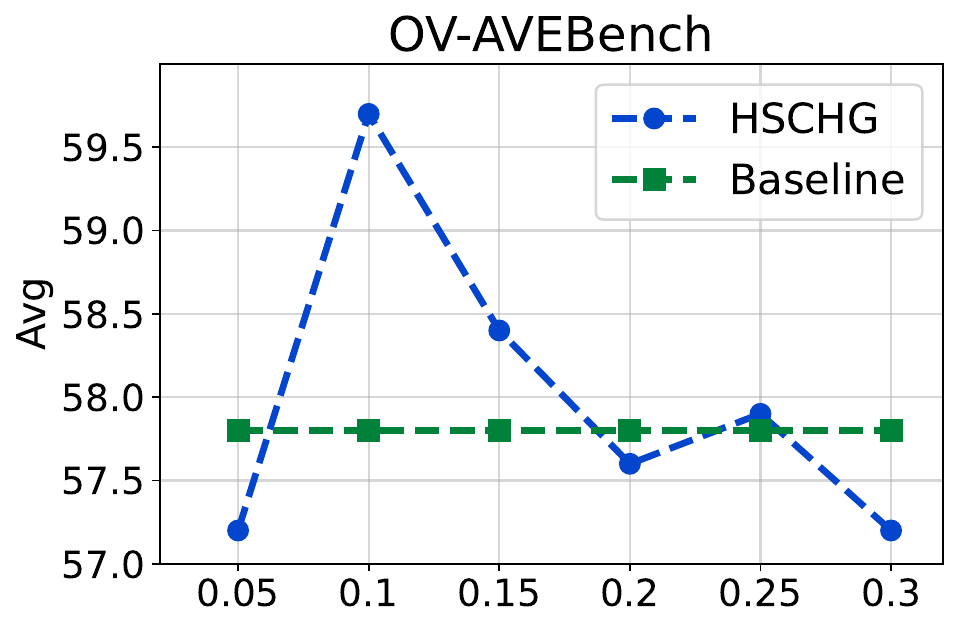}\label{gamma}}
  \hfill
  \subfloat[(b).$\tau$]{\includegraphics[width=0.48\linewidth]{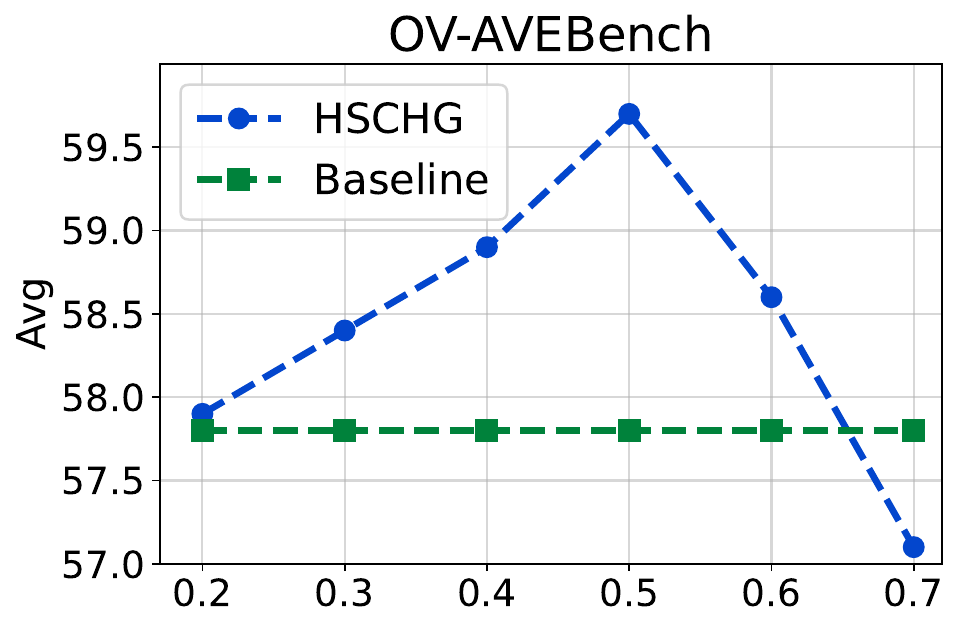}\label{tau}}\\
  \subfloat[(c).$\tau_1$]{\includegraphics[width=0.48\linewidth]{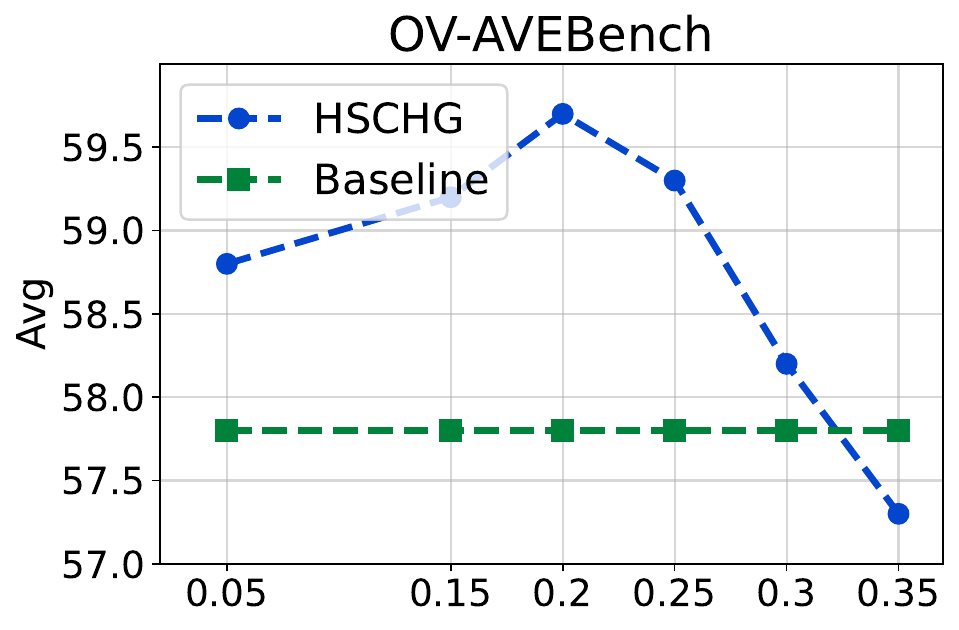}\label{tau1}}
  \hfill
  \subfloat[(d).$\tau_2$]{\includegraphics[width=0.48\linewidth]{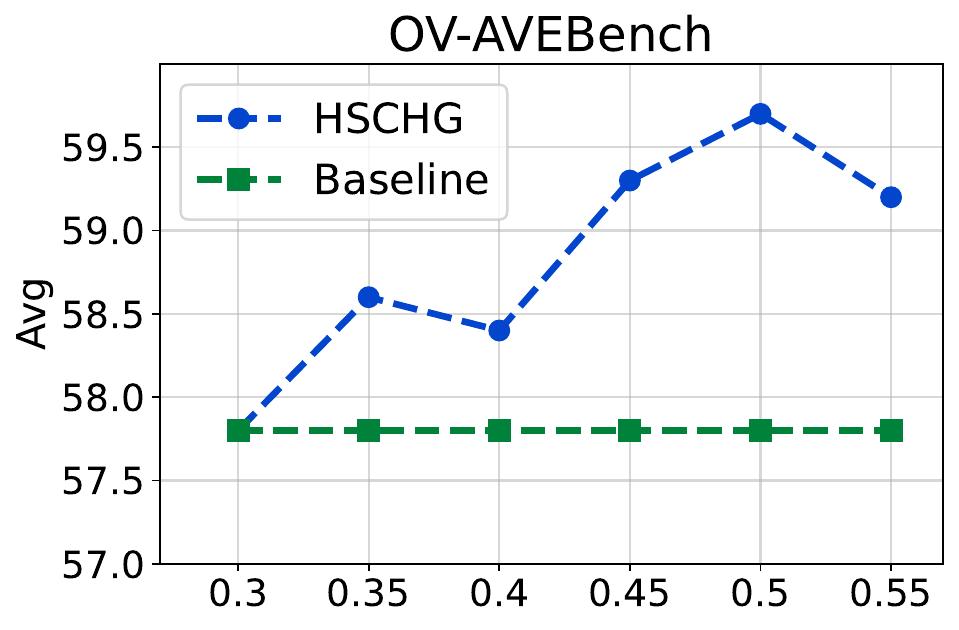}\label{tau2}}
  \caption{Hyperparameter analysis on OV-AVEBench dataset.}
  \label{O-HY}
\end{figure}

\begin{figure*}
	\centering	\includegraphics[width=0.8\linewidth]{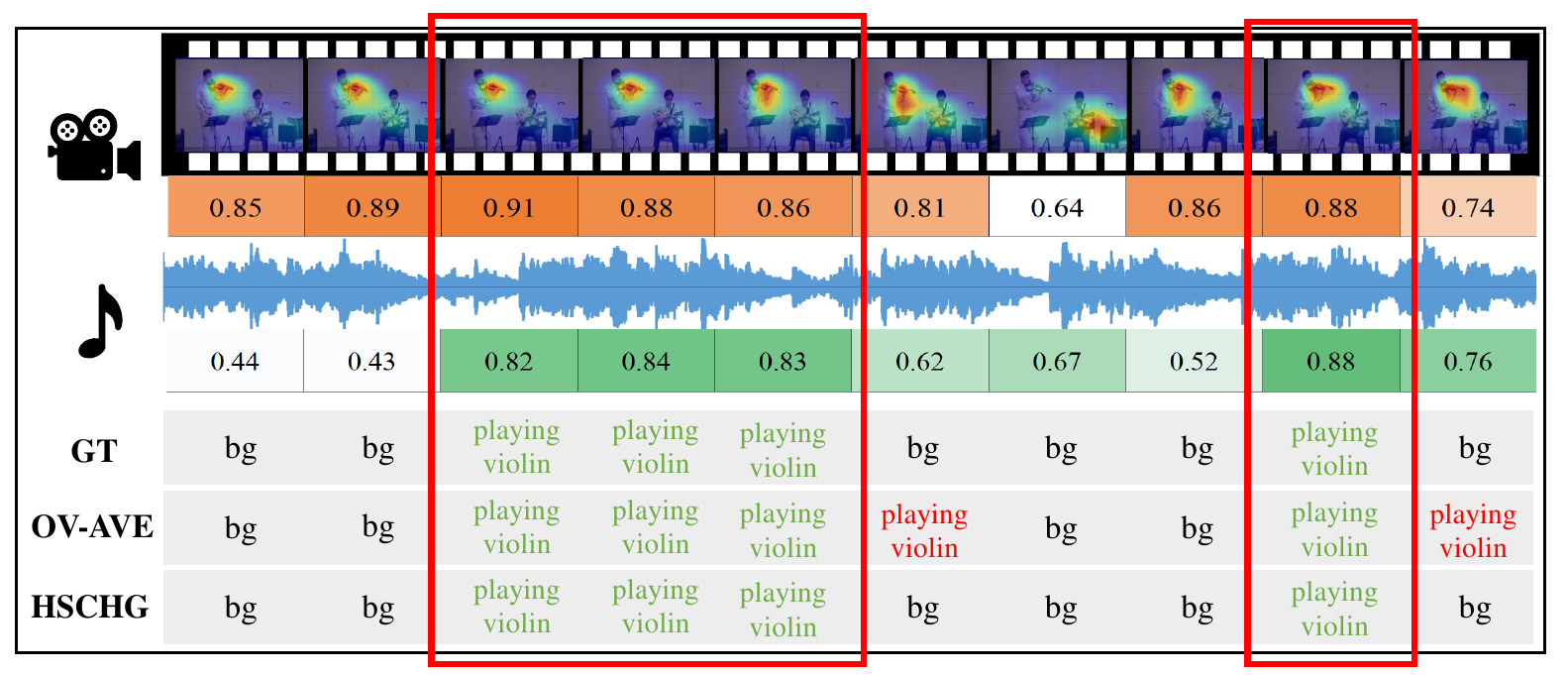}	\caption{Qualitative results of our model. The red regions stand for the answer we predict.}
	\label{reli2}
\end{figure*}

\subsubsection{Effectiveness of HHGN Components}
We further analyze the key modules of HHGN, including Multi-Directional Temporal Edges (MDTE), the Dual-Threshold Gating Mechanism (DTGM), and Bidirectional Semantic Constraints (BSC), as reported in Table \ref{HHGN}. Removing MDTE causes the largest performance drop, with Avg. decreasing to 58.5. This indicates that modeling long-range temporal dependencies is crucial for separating events from background. Without DTGM, the Avg. falls to 58.8. This confirms that the dual-threshold strategy helps filter unreliable cross-modal messages and reduces noise propagation during alignment. Ablating BSC also leads to a consistent degradation. This suggests that the closed-loop calibration between global video-level semantics and local segment-level evidence is important for robust localization.
\begin{figure}[ht]
	\centering 	\includegraphics[width=0.9\linewidth]{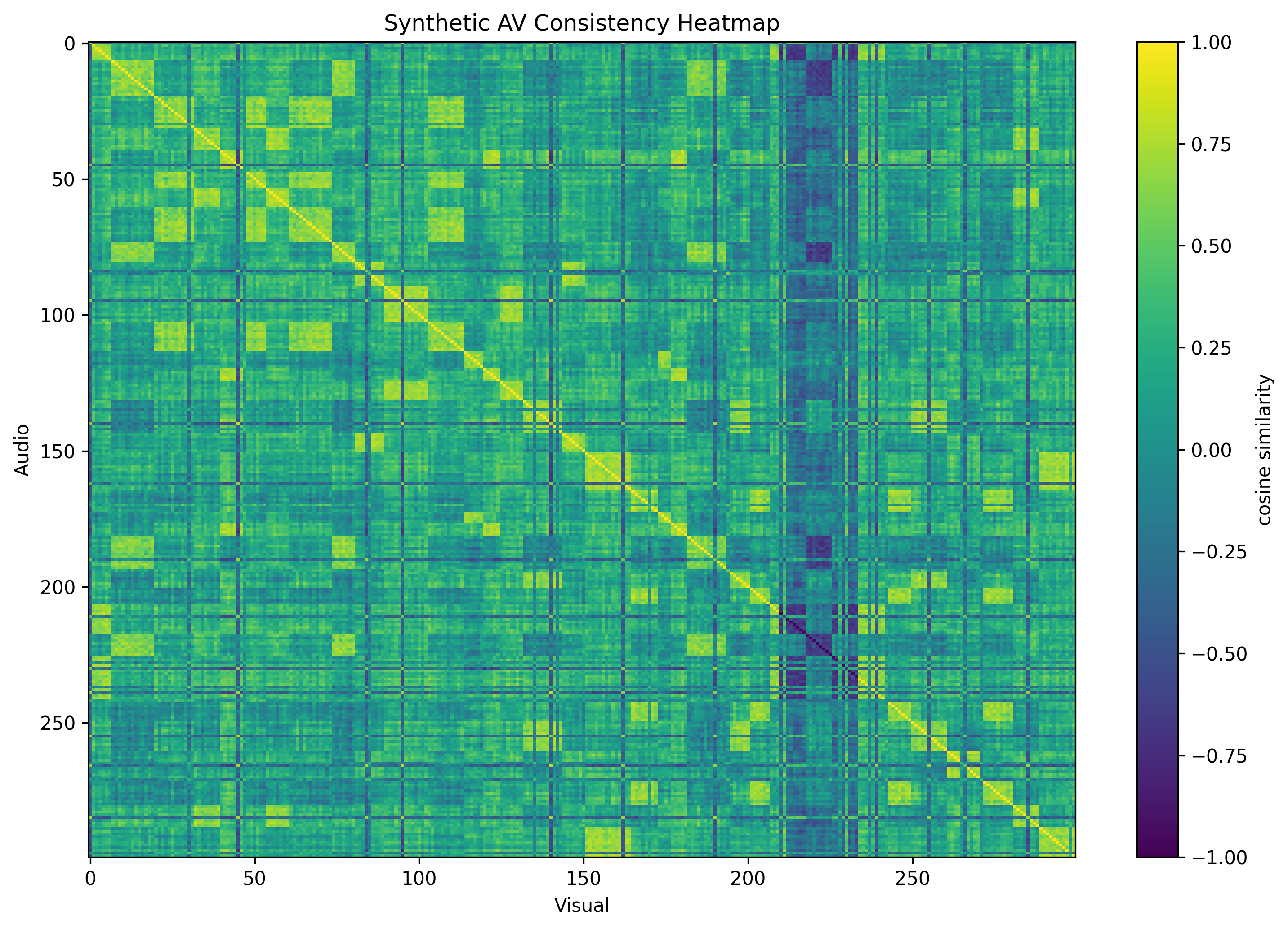}
	\caption{Audio-visual cross-modal cosine similarity heatmap: The bright main diagonal indicates temporal alignment consistency, the block-like regions correspond to similarities in repeated scenes, and the dark vertical bands and stripes reflect local mismatches or anomalous segments.}
	\label{reli1}
\end{figure}
\subsubsection{Effectiveness of Loss Functions}
Table \ref{LOSS} examines the specific contributions of the intra-modal entailment loss ($\mathcal{L}_{in}$) and the cross-modal entailment loss ($\mathcal{L}_{cr}$). Compared to using only the segmentation supervision ($\mathcal{L}_{seg}$), adding $\mathcal{L}_{in}$ improves the Avg. to 59.0, suggesting that enforcing hierarchical inclusion within modalities stabilizes feature representation. The introduction of $\mathcal{L}_{cr}$ further elevates performance (Avg. 59.1) by treating text prototypes as upper-level concepts, thereby optimizing cross-modal alignment. The full objective $\mathcal{L}_{ent}$, which combines $\mathcal{L}_{in}$ and $\mathcal{L}_{cr}$, achieves the best performance (Avg. 59.7). This confirms that intra-modal consistency and cross-modal structural alignment are complementary, jointly regularizing the embedding space to enhance both discrimination and generalization.

\subsection{Hyper Parameter Setting}

\subsubsection{Intensity Coefficient $\gamma$}
Fig. \ref{gamma} evaluates the impact of the intensity coefficient $\gamma$ on the OV-AVEBench dataset. This parameter plays a pivotal role in modulating the semantic calibration between video-level context and local segment responses. As $\gamma$ increases, the performance initially improves. This indicates that incorporating global semantic priors helps rectify local ambiguities inherent in individual segments. However, performance peaks at $\gamma=0.1$ and subsequently declines. An overly small $\gamma$ results in insufficient global guidance. This leaves local predictions vulnerable to noise. Conversely, an excessively large $\gamma$ enforces global context too aggressively. This tends to over-smooth the feature sequence and suppress discriminative local cues necessary for precise boundary detection. Consequently, we fix $\gamma=0.1$ to achieve an optimal trade-off. This setting ensures that global consistency reinforces rather than overrides local evidence.

\subsubsection{Intra-modal Threshold $\tau$}
Fig. \ref{tau} reports the sensitivity of the model to the intra-modal threshold $\tau$. This parameter governs the density of the temporal graph by filtering neighbor aggregation. In our framework, $\tau$ acts as a noise filter. It determines which neighboring segments are sufficiently relevant to contribute to the current node's representation. The results demonstrate a clear performance trajectory that peaks at $\tau=0.5$. Lower thresholds ($\tau < 0.5$) lead to the inclusion of excessive background or ambiguous segments. This dilutes the target event features with irrelevant temporal noise. On the other hand, higher thresholds ($\tau > 0.5$) yield an overly sparse neighborhood structure. This impedes the flow of temporal information and severs beneficial long-range dependencies. Thus, $\tau=0.5$ is selected to maximize the suppression of spurious connections. It also preserves the essential temporal context required for robust reasoning.

\subsubsection{Cross-modal Thresholds $\tau_1$ and $\tau_2$}
Fig. \ref{tau1} and Fig. \ref{tau2} analyze the dual thresholds $\tau_1$ and $\tau_2$. These parameters jointly define the piecewise weighting strategy for cross-modal interaction. This design is critical for handling asynchronous signals. It aims to suppress unreliable cross-modal messages when alignment confidence is low. Simultaneously, it allows effective interaction when the alignment is strong. For the lower threshold, performance improves as $\tau_1$ rises to 0.2. This confirms that filtering out weakly aligned pairs is essential to prevent the propagation of erroneous cross-modal noise. Regarding the upper threshold $\tau_2$, the model achieves optimal performance at 0.5. A value lower than this causes the model to over-activate cross-modal messages. This allows them to dominate and obscure unimodal temporal patterns. Conversely, a value set too high makes the model overly conservative. It fails to reward and utilize valid high-confidence alignments. Therefore, we set $\tau_1=0.2$ and $\tau_2=0.5$ to ensure a balanced and noise-resistant cross-modal fusion.

\begin{figure*}
	\centering	\includegraphics[width=0.9\linewidth]{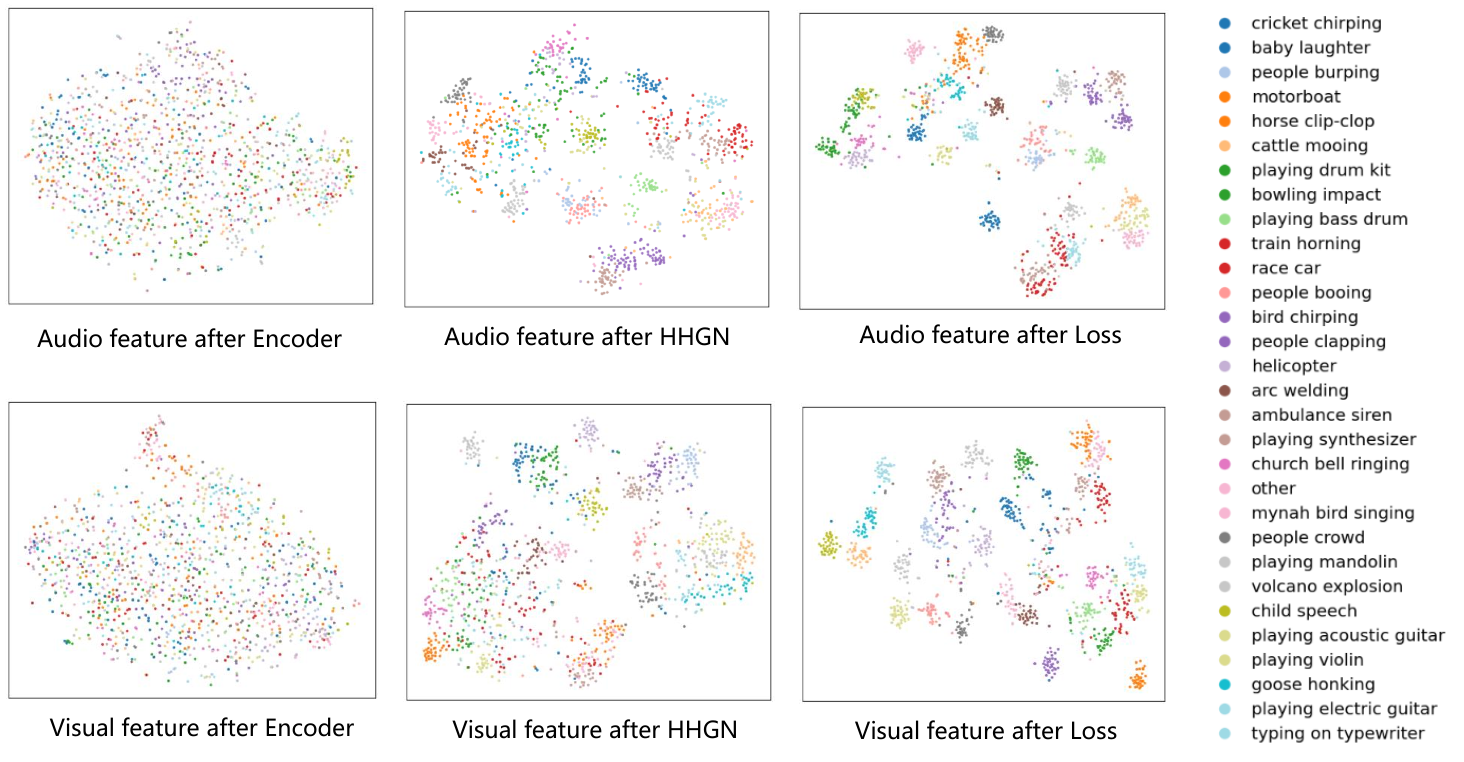}
	\caption{T-SNE visualization of audio and visual feature distributions at different stages of HSCHG.}
	\label{tsne}
\end{figure*}
\subsection{Qualitative analysis}
As shown in Fig.~\ref{reli1}, the cross-modal cosine similarity heatmap exhibits a clear enhancement along the main diagonal. This indicates that most temporal segments are synchronized across audio and visual streams. It suggests that the model can learn stable temporal alignment between the two modalities. The block-like high-similarity regions often come from repeated scenes or recurring similar segments. Such regions can introduce temporal ambiguity. Local dark vertical bands and stripes indicate cross-modal mismatches. For example, the visual signal is salient but the audio cue is missing. In other cases, the audio is strong but weakly related to the visual semantics. If these segments are directly fused, noise can be injected and propagate across time.

Fig.~\ref{reli2} further shows that our temporal-level cross-modal context suppression alleviates these inconsistencies. The figure includes the video frame response, the audio waveform, and time-varying visual and audio gating values. Within the event interval marked by the red box, both audio gating and visual response are high. The model therefore strengthens consistent cross-modal evidence in this interval. It produces a more continuous event response and matches the ground-truth boundaries better. In several background segments, the visual response remains high but the audio gate drops significantly. The model then suppresses segments dominated by unimodal cues. This helps avoid treating salient but irrelevant motion as an event. This behavior is consistent with the dual-threshold segment weighting in cross-modal interaction. Confidence filtering and gated fusion work together to reduce low-confidence message injection. As a result, boundary prediction becomes more stable and localization is more robust.

\begin{figure*}
	\centering	\includegraphics[width=0.85\linewidth]{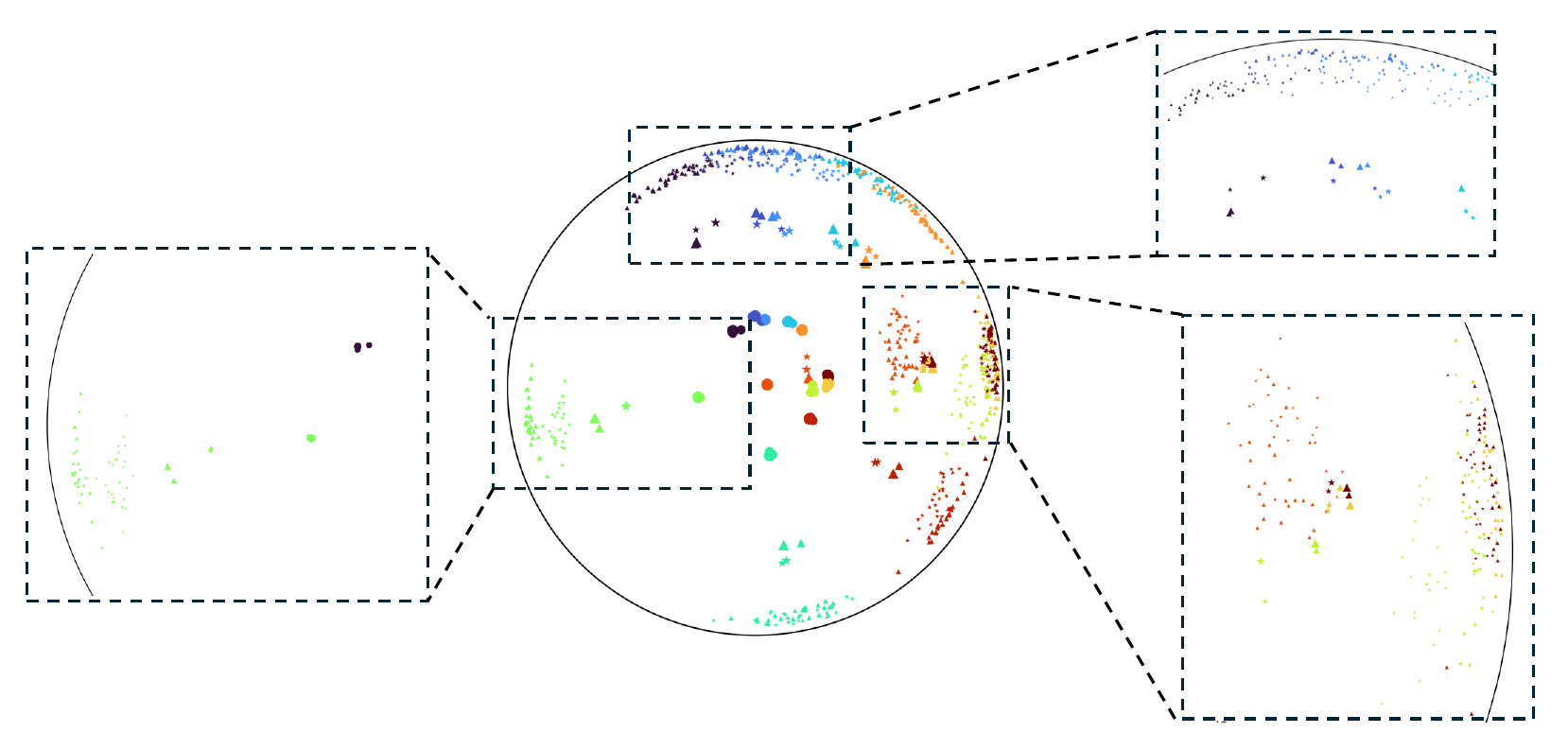}
	\caption{The HSCHG embedding is used to represent the OV-AVEBench dataset on the Poincaré disk.}
	\label{umap}
\end{figure*}
\subsection{Visualization}
\subsubsection{T-SNE Visualizations}
To further evaluate the role of HSCHG in audio-visual representation learning, we selected 30 categories from OV-AVEBench dataset and used t-SNE \cite{TSNE} to visualize the distribution of audio and video features at different stages. Fig. \ref{tsne} shows that the audio and visual features output by the encoder are severely mixed, with high intra-class dispersion and unclear inter-class boundaries, indicating that stable discriminative structures are difficult to form with only basic encoding. After HHGN, both modalities exhibit clearer cluster structures, with increased separation between categories. This indicates that HHGN improves feature consistency and robustness by performing intra-modal temporal reasoning and cross-modal interaction on a hierarchical heterogeneous graph containing segment nodes and video nodes. After adding the hierarchical entailment loss, the clusters become tighter and more distinctly separated, and the categorical organization of audio and visual features becomes more consistent. This demonstrates that hierarchical constraints between and within modalities can effectively improve the model's generalization ability under open-vocabulary conditions.

\subsubsection{UMAP Visualizations}
To demonstrate the advantage of hyperbolic embedding optimization in HSCHG for hierarchical semantic modeling, inspired by \cite{AAAI}, we use UMAP \cite{UMAP} to project the optimized embeddings into a 2D space, as shown in Fig. \ref{umap}. Different colors denote different classes. Circles represent text prototypes, triangles represent audio features, and pentagrams represent visual features. Point size indicates feature granularity. Larger points correspond to video-level representations, while smaller points correspond to segment-level representations. We can see that the negative curvature of the hyperbolic space helps represent the hierarchical structure of audio-visual semantics. Video-level features tend to lie closer to the inner region, while segment-level features are more concentrated in the outer region and form local groups around their corresponding video-level features. This pattern matches the semantic inclusion relation, where segment semantics gradually aggregate into video-level semantics, which supports the ability of hyperbolic space to model tree-like or hierarchical structures. In addition, text prototypes often appear near the center or the main structure of each class cluster, and audio and visual samples lie nearby. This indicates strong cross-modal consistency in a shared space. Compared with Euclidean space, hyperbolic space can separate different levels more clearly in limited dimensions. Some semantically related classes are closer in local regions, which suggests that the model captures shared structure in the high-dimensional space. A small number of samples are close across classes. These often correspond to basic segments or background cues with common audio-visual patterns. They remain consistent within a class while also showing shared traits across classes. This indicates that the model learns a hierarchy-aware organization rather than simple class separation. This visualization supports our claim that hyperbolic embedding optimization can better capture hierarchical relations in audio-visual data and improves generalization in the open-vocabulary setting.

 

\section{Conclusion}
\label{D}
This paper introduces HSCHG, a hierarchical semantic constrained heterogeneous graph framework for open vocabulary audio-visual event localization. HSCHG first builds a heterogeneous hierarchical graph in Euclidean space to model segment- and video-level audio-visual semantics. It learns audio-visual consistency through intra-modal temporal edges, cross-modal edges, and hierarchical edges. HSCHG then maps the learned multi-level representations into hyperbolic space. It enforces entailment based constraints with text prototypes to preserve hierarchical relations and support open vocabulary generalization. Experiments on the OV-AVEBench dataset show consistent gains in both localization and recognition, indicating improved robustness and generalization.

\bibliographystyle{unsrt}
\bibliography{refs}

\vfill

\end{document}